\documentclass[10pt,twocolumn,letterpaper]{article}
\usepackage{cvpr}
\usepackage{times}
\usepackage{epsfig}
\usepackage{graphicx}
\usepackage{amsmath}
\usepackage{amssymb}
\usepackage{multirow}
\usepackage{rotating}
\usepackage[noadjust]{cite} 
 \usepackage[none]{hyphenat}
\usepackage{flushend}
\usepackage[pagebackref=true,breaklinks=true,letterpaper=true,colorlinks,bookmarks=false]{hyperref}

\begin{document}

\title{Deep Generative Attacks and Countermeasures for \\ Data-Driven Offline Signature Verification}

\author{An Ngo \\
Bucknell University, USA\\
{\tt\small axn001@bucknell.edu}
\and
Rajesh Kumar\\
Bucknell University, USA\\
{\tt\small rajesh.kumar@bucknell.edu}
\and
Phuong Cao\\
Bucknell University, USA\\
{\tt\small mtc013@bucknell.edu}
}

\maketitle
\thispagestyle{empty}

\begin{abstract}
This study investigates the vulnerabilities of data-driven offline signature verification (DASV) systems to generative attacks and proposes robust countermeasures. Specifically, we explore the efficacy of Variational Autoencoders (VAEs) and Conditional Generative Adversarial Networks (CGANs) in creating deceptive signatures that challenge DASV systems. Using the Structural Similarity Index (SSIM) to evaluate the quality of forged signatures, we assess their impact on DASV systems built with Xception, ResNet152V2, and DenseNet201 architectures. Initial results showed False Accept Rates (FARs) ranging from $0\%$ to $5.47\%$ across all models and datasets. However, exposure to synthetic signatures significantly increased FARs, with rates ranging from $19.12\%$ to $61.64\%$. The proposed countermeasure, i.e., retraining the models with real \textit{+} synthetic datasets, was very effective, reducing FARs between $0\%$ and $0.99\%$. These findings emphasize the necessity of investigating vulnerabilities in security systems like DASV and reinforce the role of generative methods in enhancing the security of data-driven systems.

\end{abstract}

\section{Introduction}
Handwritten signatures have long been a cornerstone of identity verification, utilized extensively in legal, financial, and medical domains. Despite their widespread use, the manual verification of signature authenticity remains a laborious and impractical process, especially at large scales. The rise in sophisticated signature frauds further complicates manual verification, underscoring the need for automated solutions. Consequently, Automated Signature Verification (ASV) has been a vibrant field of research for several decades \cite{OfflineASVReview, MLBasedOFFSign2021Survey}. Traditional machine learning classifiers, such as Hidden Markov Models, Support Vector Machines, and Neural Networks, have been employed with notable success, achieving commendable error rates \cite{OfflineASVReview, MLBasedOFFSign2021Survey}. More recently, the advent of data-driven approaches, particularly Convolutional Neural Networks (CNNs) and Transformer-based architectures, has significantly enhanced representation learning and improved verification accuracy \cite{DeepCNN2017, StatsBased2017, MLBasedOFFSign2021Survey, StaticSignatureGeneration2021, TransformerOffline1, TransformerOffline2}. CNN-based models, in particular, have demonstrated lower error rates across diverse datasets, affirming their scalability and generalizability \cite{DeepCNN2017, cnnacc}.

The threat of signature forgery poses substantial risks to various sectors, potentially leading to severe financial losses, legal disputes, and breaches of confidential information. These risks highlight the critical need for robust ASV systems to resist sophisticated forgeries. The literature differentiates between writer-dependent and writer-independent models within the domain of ASV. This study focuses on writer-dependent models, which are meticulously tailored to individual users \cite{OfflineASVReview}. Furthermore, ASV systems are classified based on data acquisition techniques into online and offline verification systems. This paper concentrates on offline ASV systems, which rely on static signature images while deferring the exploration of online systems to future research \cite{Survey2019, Attack2023Survey, MLBasedOFFSign2021Survey, OfflineASVReview}.

Like other biometric systems, ASVs are vulnerable to various types of attacks, which can be categorized as either random or skilled based on the attacker's knowledge and intent \cite{Attack2023Survey}. Random forgeries are executed without prior knowledge of the target signature. In contrast, skilled forgeries involve deliberate attempts to replicate or mimic the target signature with varying degrees of sophistication, from rudimentary imitations to highly practiced and refined copies \cite{Forgeries2018}. The increasing complexity of these attacks necessitates the development of innovative defensive measures.

Ballard et al. \cite{BallardAttackCategory2007} expanded the classification of attacks to include naive, trained, and generative methods, highlighting the critical role of generative attacks in evaluating ASV security. While there has been some research into the generation of synthetic signatures \cite{OnlineGenerationiDeLog2018, StaticSignatureGeneration2021}, the potential of advanced generative techniques, such as Generative Adversarial Networks (GANs) and Variational Autoencoders (VAEs), remains largely unexplored \cite{goodfellow2014generative, KingmaWellingVAE}. Therefore, this study explores the application of contemporary Deep Generative Models (DGMs) to challenge data-driven ASV systems and develop effective countermeasures. The primary contributions are as follows:

\begin{itemize}
    \item Utilized Variational Autoencoders (VAE) \cite{KingmaWellingVAE} and Conditional Generative Adversarial Networks (CGAN) \cite{MirzaCGAN} to generate six synthetic datasets from forgeries in CEDAR \cite{CEDARDataset}, BHSig260-B \cite{Pal2016BHSig}, and BHSig260-H \cite{Pal2016BHSig}, and evaluated their quality using the Structural Similarity Index (SSIM) \cite{Wang2004SSI}.
    \item Evaluated the robustness of DenseNet201 \cite{DenseNet2017}, ResNet152V2 \cite{ResNet2016}, and Xception \cite{Xception} architectures against various attack methods.
    \item Identified a correlation between the SSIM of the generated images and the attack success rates measured by False Accept Rates (FAR), thereby introducing controlled and explainable attack mechanisms.
    \item Enhanced model performance by incorporating SSIM-optimized synthetic forgeries into the training process. 
\end{itemize}

The remainder of the paper is organized as follows: Section \ref{RelatedWork} provides a review of the relevant literature, Section \ref{ExperimentalSetup} details the experimental setup, Section \ref{ResultsAndDiscussion} presents the findings and discussion, and Section \ref{ConclusionFutureWork} concludes the study and outlines future research directions.

\section{Related work}
\label{RelatedWork}
Previous studies explored the use of deep convolutional neural networks (CNNs) with Siamese architectures for offline signature verification \cite{DeepCNN2017, signet2017}. These studies showed improvements in verification performance across various datasets, including reduced error rates on the CEDAR dataset. Other architectures, such as Inception \cite{Inception2015}, VGG16 \cite{VGG2014}, and DenseNet121 \cite{DeepSign2021AllBaselines}, have also been used effectively. Yusnur et al. \cite{ResNetDenseNet2022} implemented CNNs with triplet loss, focusing on ResNet-18 and DenseNet-121. In this study, we use the latest versions of these architectures to establish baseline models.

Besides signature verification, automatic signature generation has been researched for several years. Ferrer et al. \cite{SignatureGeneration2013, StaticSignatureGeneration2014} focused on generating forgeries and new identity signatures, developing a two-stage method mimicking the human handwriting process. Diaz et al. \cite{FirstGeneration2017} introduced a cognitive-inspired algorithm that accounts for human spatial and motor variabilities in signing. Maruyama et al. \cite{StaticSignatureGeneration2021} proposed a method to capture common writer variability traits and introduced a way to assess the quality of generated signatures using feature vectors. Further details of applications of signature generation can be found in \cite{FirstGeneration2017}.

Recent surveys have highlighted the threat posed by generated adversarial examples in offline signature verification and developing countermeasures \cite{MLBasedOFFSign2021Survey}. For instance, Hafemann et al. \cite{hafemann2019characterizing} identified vulnerabilities in CNN-based verifiers to deceptive inputs. Szegedy et al. \cite{adversarialidea} noted the susceptibility of neural network-based ASVs to perturbed signatures. Bird \cite{bird2022robotic} explored synthetic forgeries generated using Conditional GANs and robotic arms, documenting false acceptance rates and recommending system training with synthetic forgeries as a countermeasure. In contrast, Li et al. \cite{li2021} suggested that while adversarial inputs are detectable by humans, they failed to deceive ASVs. Tolosana et al. \cite{VAEForGeneration} investigated the use of Variational Autoencoders (VAEs) for \textit{online} handwriting synthesis but did not evaluate the threat they posed to ASVs. In contrast, our research focuses on \textit{offline} signature verification, providing further insights into the understanding of generative adversarial threats. In particular, we employ state-of-the-art generators such as Conditional GANs and VAEs to challenge baseline models implemented using Xception, DenseNet201, and ResNet152V2—across three different datasets, each in a different language. We also emphasized controlling the quality of synthetic signatures to devise effective countermeasures, as detailed in Section \ref{generate_method}.

The presented attack scenarios simulate black-box injection attacks to evaluate ASV vulnerabilities, keeping real-world scenarios in mind where attackers do not have detailed knowledge of the baseline models. Please see Section \ref{threatmodel} for more details.  

\section{Design of experiments}
\label{ExperimentalSetup}
\begin{figure}[ht]
  \centering
  \includegraphics[width=3.2in, height = 1.0in]{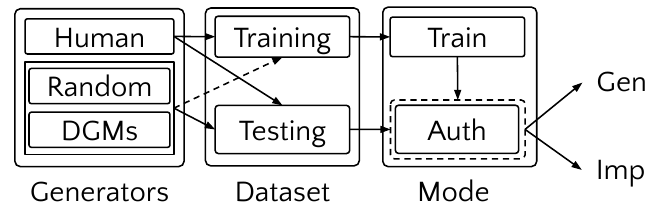}
  \caption{Illustrating the process from data generation to model evaluation. Data sources include human-generated signatures, random noise, and deep generative models (DGMs), creating training and testing datasets. During training, baseline models are trained. In testing, models are first evaluated on accepting genuine and rejecting forged signatures, then tested for susceptibility to generated signatures. Models are then retrained with generated samples and re-evaluated for both criteria. Performance is measured using false reject rates (FRR) and false accept rates (FAR). Solid arrows represent baseline training, while dashed arrows indicate retraining.}
  \label{SAAuthPipeline}
\end{figure}

Figure \ref{SAAuthPipeline} provides an overview of the experimental setup. The components and corresponding sub-components are described in the subsequent paragraphs. The dataset and implementation details are available via Github\footnote{https://github.com/axn0684/DeepGenerativeAttack}

\subsection{Datasets}
\label{datasets}
To test our attack and countermeasure approaches, we used three different datasets, each consisting of signatures from a different language: 

\textbf{CEDAR} This widely used dataset in the offline ASV domain \cite{CEDARDataset} includes $24$ genuine and $24$ forged signatures for each of the $55$ users, totaling $1320$ genuine and $1320$ forged signatures. The signatures were at $300$ dpi gray-scale and were binarized using a gray-scale histogram. Salt-and-pepper noise removal and slant normalization were also applied to preprocess the signature images.

\textbf{BHSig260Bengali} This dataset consists of $100$ sets of handwritten offline signatures. Each set comprises $24$ genuine and $30$ skilled forgeries. All signatures were collected using a flatbed scanner with a resolution of $300$ dpi in gray-scale and stored in TIFF format. A histogram-based threshold was applied for binarization to convert data to two-tone images. The Bengali script has more curved shapes compared to Hindi Devanagari. It also includes unique nasal consonants and sibilants not found in Hindi, while vowel forms have a horizontal orientation.

\textbf{BHSig260Hindi} This dataset consists of $160$ sets of handwritten offline signatures. The components and acquisition of each set are the same as the BHSig260-Bengali dataset. The angular Devanagari script is used for Hindi signatures. It contains distinct consonant clusters that are absent in Bengali. The Hindi vowel shapes have a more diagonal and vertical orientation than Bengali. While both datasets contain Indian script signatures, the writing styles differ due to variations in the Bengali and Hindi character sets.
 
\textbf{Random signatures} To train the ASVs, Otsu preprocessing and normalization proposed in \cite{OtsuPreprocessing}, were applied on the original datasets besides transforming the signatures into a binary array of size $224$x$224$. Thus, to generate random signatures, we created arrays of size $224 \times 224$ filled with randomly selected values of either $0$ or $1$. These random signatures were used to launch random signature attacks and evaluate ASVs' vulnerability to such attacks.

\textbf{DGM-assisted synthetic signatures} We used VAE \cite{VAEForGeneration} and CGAN \cite{MirzaCGAN} to create six synthetic datasets from the forgeries of each public dataset. Our generation method differs from that of Bird et al. \cite{bird2022robotic}, who supplied conditional GANs with multiple images labeled \textit{genuine} and \textit{forgery} to generate GAN-based images. We trained individual models for each generative architecture and generated nine synthetic samples for each original image. The six synthetic datasets were divided into two parts, training, and testing, to avoid data leakage. The training part comprised signatures generated from references drawn from the training set of human forgeries, while the testing part comprised signatures generated from references drawn from the testing set of human forgeries. We describe the generators below. 

\textbf{CGANs} are an extension of the vanilla GAN \cite{goodfellow2014generative}. They consist of two components: a generator and a discriminator. The generator creates synthetic data samples, while the discriminator distinguishes between real and generated data. The key difference in CGANs is the introduction of additional conditional variables, such as labels or tags, to both the generator and the discriminator. This conditional information provides control over the types of data the generator creates. In this study, CGANs were trained on existing datasets to create synthetic forgeries of each public dataset. Mathematically, the objective function of a CGAN is expressed as:
\[
    \min_G \max_D V(D, G) = \mathbb{E}[\log D(x|y)] + \mathbb{E}[\log(1 - D(G(z|y)))]
\]

\textbf{VAEs} are generative models consisting of two main parts: an encoder and a decoder. The encoder compresses input data, in this case, images, into a lower-dimensional latent space while the decoder reconstructs the original data from these compressed representations. VAEs use a variational inference approach, introducing a probabilistic layer that models the latent space as a distribution, typically Gaussian, rather than deterministic values. In this study, VAEs were trained on images from three human forgery datasets. The encoder was designed to map signature images to a lower-dimensional latent space, capturing the essential features and variations of the signatures. The decoder was trained to generate signature images from the latent representations.

Mathematically, VAEs optimize an objective function that includes a reconstruction loss and a regularization term. The objective function of VAE is expressed as:
\[
\mathcal{L}(\theta, \phi; x) = -\mathbb{E}[\log p_\theta(x|z)] + \text{KL}(q_\phi(z|x) || p(z))
\]
Here, $x$ represents the input data, which are human forgeries, $z$ is a sample from the latent space, $q_\theta$ is the encoder, $p_\theta$ is the decoder, and KL represents the Kullback-Leibler divergence.


\textbf{Quality of synthetic signatures} was evaluated using the Structural Similarity Index Measure (SSIM) \cite{Wang2004SSI}. SSIM considers luminance (brightness), contrast, and the structure of pixels in the images. It calculates three key components: luminance ($l$), contrast ($c$), and structure ($s$), each designed to capture specific aspects of similarity. These components are computed using mean ($\mu$) and standard deviation ($\sigma$) values of pixel intensities within the windows. The SSIM score, denoted by $d(x,y)$, is computed as follows:

\[
l(x, y) = \frac{2 \cdot \mu_x \cdot \mu_y + C_1}{\mu_x^2 + \mu_y^2 + C_1}
\]

\[
c(x, y) = \frac{2 \cdot \sigma_x \cdot \sigma_y + C_2}{\sigma_x^2 + \sigma_y^2 + C_2}
\]

\[
s(x, y) = \frac{\sigma_{xy} + C_3}{\sigma_x \cdot \sigma_y + C_3}
\]

\[
d(x, y) = l(x, y) \cdot c(x, y) \cdot s(x, y)
\]

The value of $d(x, y)$ ranges from $-1$ to $1$, where $1$ indicates perfect structural similarity, $0$ denotes no similarity, and negative values indicate dissimilarity.

VAE-generated datasets achieved an average $d(x, y)$ of $0.8$ across three datasets. In contrast, the average $d(x, y)$ for CGAN-generated datasets was $0.11$. Details are presented and discussed in Table \ref{SSITable}.

One could argue that CGAN should achieve better results than VAE, in general while generating the synthetic signatures. However, this was not the case in our experimental setup, likely because (1) we are generating new signatures using a single reference signature, and (2) GAN suffers from limited data for training, mainly because the discriminator memorizes the training set \cite{LimitedDataGAN}, and (3) VAE has inherent regularization in the VAE's objective (the KL divergence term)-- preventing over-fitting to the single reference sample, presumably a more stable training process, and the smooth latent space it learns.

\textbf{SSIMs vs. FARs (attack success)} We used a fixed number of epochs ($800$) to generate the synthetic signatures, and the resulting $d(x, y)$ values are presented in Table \ref{SSITable}. VAE achieved average SSIMs of $0.942$, $0.510$, and $0.938$ for the Bengali, CEDAR, and Hindi datasets, respectively. With the same number of epochs, CGAN achieved average SSIMs of $0.104$, $0.126$, and $0.108$ for the Bengali, CEDAR, and Hindi datasets, respectively.

\begin{figure}[htp]
  \centering
  \includegraphics[width=3.3in, height=1.9in]{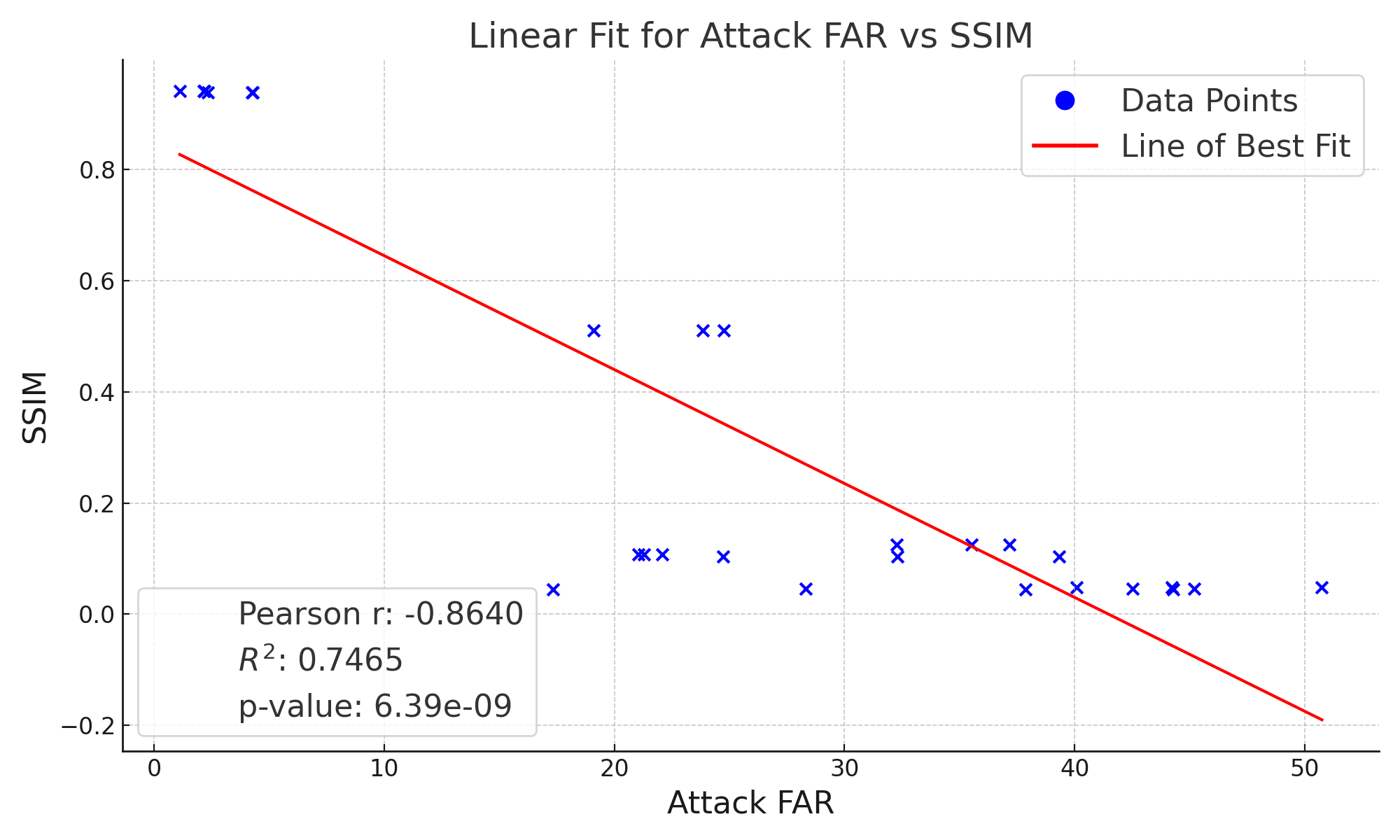}
  \caption{The Pearson correlation coefficient ($r$) of $-0.8640$ and $R^2$ of $0.7465$ demonstrate a strong negative correlation between SSIM and FAR, with a p-value of 6.39e-09 confirming statistical significance. This means lower-quality synthetic forgeries (lower SSIM) are more likely to be accepted as genuine by the verification system because the generated samples are farther from the forgeries and, in turn, closer to genuine signatures in a two-class setup. This observation formed the basis for a novel countermeasure, i.e., retraining models with SSIM-controlled datasets, to enhance the robustness of DASV systems against synthetic forgeries.}
  \label{FAR SSI}
\end{figure}

Observing the stark difference between the damages (FARs) caused by the VAE-generated and CGAN-generated signatures (see Figure \ref{attack-baseline}), we investigated the relationship between $d(x, y)$ and the false accept rates achieved by the generated signatures. A strong negative correlation was found between $d(x, y)$ and the false accept rates, as demonstrated in Figure \ref{FAR SSI}, which offers insights into controlling the quality of the generated signatures and, in turn, the ability to attack.

The uncovered relationship between SSIM and FARs led us to generate an SSIM-controlled dataset for effective attacks and developing countermeasures. Since attackers can always obtain impostor signature samples, we centered the data generation process around impostor samples. In other words, we used impostor samples as the reference for data generation via DGMs.

\textbf{SSIM-controlled generation}
\label{generate_method} Our objective was to generate signatures that deviate significantly from the impostor samples so they could be misclassified as genuine, thereby increasing the FARs. Specifically, we aimed to generate signature samples from forgeries with the lowest SSIM.

We generated five signatures per user from the impostor samples at different epochs. Upon plotting the results, we observed that CGAN could only generate signatures with very low SSIM. In contrast, VAE could generate signatures with a wide range of SSIMs (see Table \ref{SSITable} and Figure \ref{fig:dgm-ssim-distribution}).

\begin{table}[htbp]
\caption{Comparison of SSIM values generated by VAE and CGAN for three datasets (Bengali, CEDAR, and Hindi). The heat distribution of SSIM achieved by CGAN and VAE-based signature generators for different epochs is shown in Figure \ref{fig:dgm-ssim-distribution}. VAE was selected for further tuning to produce low SSIM images required for retraining the authentication models. This selection was due to VAE's ability to generate a wider range of SSIM values, including very low SSIMs necessary to exploit the negative correlation between SSIM and FAR. CGAN could not match the performance of VAE, even with a high number of epochs. Higher SSIM values indicate a better match with reference images, while lower SSIM values indicate a greater deviation.}
\vspace{0.1in}
\label{SSITable}
\centering
\begin{tabular}{lrrr}
\hline
Dataset   &   VAE&   CGAN&   VAE SSI-tuned\\
\hline
Bengali   &      0.942 &       0.104 &               0.048 \\
CEDAR     &      0.510 &       0.126 &               0.045 \\
Hindi     &      0.938 &       0.108 &               0.045 \\
\hline
\end{tabular}
\end{table}
Considering the high SSIM images generated by VAE compared to CGAN, we decided to retrain VAE with the SSIM-controlled countermeasure (see Table \ref{SSITable}). Using this approach to attack the baselines, we generated signatures with SSIM less than $0.05$ to validate our observation of the relationship between SSIM and FAR.

The VAE models were retrained with the same dataset until the generated signatures reached an SSIM that could be considered insignificant. Specifically, instead of training our deep-learning baselines for a fixed number of $800$ epochs and using the final generator to produce signatures, we sampled one signature every $5$ epochs during the training. If the signatures were within the threshold (in this case, $0.04-0.08$), we stopped the training and kept the sampled signature. If, after $800$ epochs, no signature met the criteria, the model was retrained from scratch. This process was repeated $9$ times for each signature to obtain $9$ generated samples, similar to the uncontrolled dataset. We named this dataset the VAE-SSI-Controlled dataset.

The new VAE-SSI-Controlled dataset further confirmed our observation of a negative correlation between SSIM and FAR, as it achieved higher FARs than the epoch-guided ones. The epoch-guided dataset is referred to as the VAE dataset, while the SSI-controlled dataset will be referred to as VAE-SSI-Controlled hereafter.

\subsection{Baseline implementation}
\label{baselines_sec}
The baselines were created using three advanced Deep Convolutional Neural Network architectures: Xception, ResNet152V2, and DenseNet201. These architectures were chosen primarily because their previous versions demonstrated effectiveness in the literature for signature verification \cite{InceptionSigver,ResNetSigVer,DenseNet2017}. They mainly served as feature extractors to train user-specific signature authentication models. Since the focus is on writer-dependent models, a baseline was created for each user using their signatures.

\textbf{Preprocessing} We applied the Otsu method \cite{OtsuPreprocessing} for signature preprocessing. Subsequently, the images were resized to $224 \times 224$ using the bi-cubic interpolation method and normalized before being input for feature extraction. The same preprocessing technique was also applied during testing and attacking.

\textbf{Training and testing the authentication models} We trained individual models for each user using two-thirds of their genuine and forged signatures. The remaining one-third of each user's dataset and the synthetic signatures generated from their genuine and forged signatures were used for testing. The authentication module consists of a simple dense layer with $256$ nodes using the ReLU activation function and a dense layer with two nodes representing the output layer for genuine and impostor signatures. These layers use the Sigmoid activation function. We employed binary cross-entropy with a Stochastic Gradient Descent (SGD) optimizer and a learning rate of $0.0001$ with momentum of $0.9$. The resulting models are referred to as Vanilla Baselines to distinguish them from those obtained after retraining.

\subsection{Threat model}
\label{threatmodel}
This paper considers black-box attacks, similar to those presented in \cite{Zhao2020DataInjection} for deceiving gait, touch, face, and voice-based verification. The attackers do not require access to the details of the baseline models. However, the attackers need access to the samples the verifiers accept. Legitimate samples can be obtained through theft or using the membership inference technique presented in \cite{InferenceAttack}.

\begin{figure}[htp]
    \centering
    \vspace{-0.1in}
    \includegraphics[height =2.6in, width=3.2in]{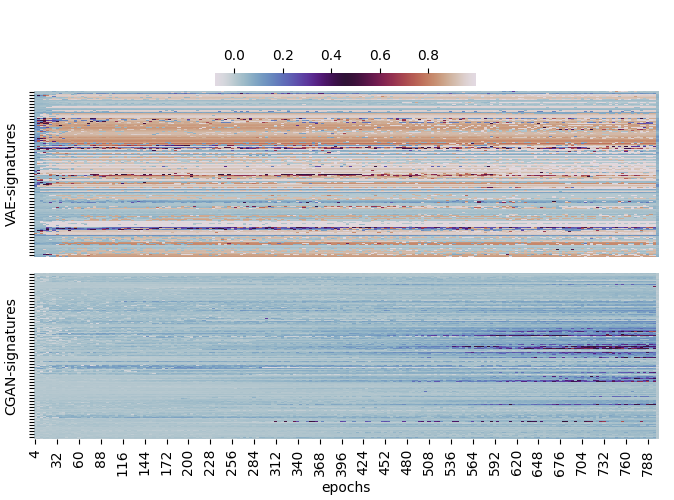}
    \caption{The heatmap of SSIMs achieved by CGAN and VAE-based signature generators across different epochs. On the y-axis are 275 signature samples generated by each generator, with five samples per user. VAE could generate signatures with a wide range of SSIMs, i.e., signatures close to and far from the reference signatures (skilled forgeries) across varying epochs. In contrast, CGAN could not match VAE's capability to generate a variety of signatures with differing SSIMs. In other words, VAE outperformed CGAN in controlled signature generation, thus chosen for further experiments.}
    \vspace{-0.2in}
    \label{fig:dgm-ssim-distribution}
\end{figure}

\subsection{Attack evaluation scenarios}
We used four attack scenarios to assess the baseline models' attack susceptibility. The impacts of each attack were measured using False Acceptance Rates (FARs). The higher the FAR, the more successful the attack.

\textbf{Vanilla or zero-effort attack}
The Vanilla Attack involves traditional impostor testing with human forgeries in the corresponding datasets. This attack is the primary scenario used to evaluate baseline performance, facilitate comparison with previous studies, and highlight the performance of various other attacks.

\textbf{Random attack}
We adopted this attack from \cite{Zhao2020DataInjection, gaitgan}, which studied random vector attacks for deceiving gait, touch, face, and voice-based verification. Following this approach, we generated $5000$ random images to attack and evaluate each baseline.

\textbf{VAE and CGAN attack}
In this scenario, we used VAE and CGAN-generated signatures to test if they could bypass the signature verification process. Unlike VAE-generated signatures, CGAN-generated signatures had more significant disparities from the reference images (see Figure \ref{fig:dgm-ssim-distribution}). We hypothesized and expected CGAN-generated signatures to produce a higher FAR because the generated images were farther from the impostor images, resulting in a higher chance of being classified as genuine. This hypothesis was further investigated by exploring the relationship between the similarity of the generated and original images and the corresponding false accept rates (FAR).

\subsection{Countermeasures} 
\label{retrain_process}
\textbf{Retraining with synthetic datasets}
We propose synthetic data-augmented retraining of DASVs to enhance their robustness against the studied attacks. Each DASV was retrained using random signatures, DGM-assisted signatures, and human forgeries as impostor samples. This process resulted in three new models: random-assisted, VAE-assisted, and CGAN-assisted. These retrained models were then evaluated using the same attack scenarios and metrics. Separate datasets were used for retraining and testing to avoid data leakage.

For random-assisted systems, the random signatures were used in retraining. Similarly, the VAE and CGAN datasets were divided into training and testing sets for synthetic dataset-assisted retraining. We refer to the retrained models as VAE-assisted and CGAN-assisted verification models. These models were tested on synthetic signatures from the testing datasets to ensure the retraining set remained unseen. This process aimed to improve system performance in unfamiliar contexts.

\textbf{SSIM-Tuning}
While evaluating the baseline models under GAN-assisted attack scenarios, we observed a strong negative correlation between SSIM and FAR (specified in Section \ref{Attack_Per}). In other words, lower SSIM values (indicating greater distance from the reference human forgeries) corresponded to higher FARs, as these signatures were more likely to be accepted as genuine. Based on this observation, we posit that training models with low SSIM signatures, considering them as impostors would increase impostor sample variance, reducing FAR and enhancing robustness against various attack scenarios.

We decided to generate SSIM-controlled signatures for retraining the baselines. Among CGAN and VAE, VAE was more suitable for SSIM-controlled tasks due to its ability to generate images with a wide range of SSIMs (see Figure \ref{fig:dgm-ssim-distribution}). The new VAE models were trained with the same dataset settings until the generated images reached an insignificant SSIM. Instead of training our deep-learning baselines with a fixed number of $800$ epochs, we sampled one image every $5$ epochs. If the image was within the threshold $(0.04-0.08)$, we stopped the training and kept the sampled image. The model was retrained from scratch if no image met the criteria after $800$ epochs. This process was repeated $9$ times for each signature image to obtain $9$ generated samples, similar to the uncontrolled dataset.

It took several iterations to generate images with the desired SSIM. However, the process did converge in a certain number of iterations. This process is a bit ad-hoc, therefore in the future, we will investigate if it can be simplified further.  

The new VAE SSI-controlled dataset further confirmed our observation of a negative correlation between SSIM and FAR. The new VAE SSI dataset posed a greater threat to the baseline than the initial VAE dataset and was used to test or retrain the models. This paper refers to the VAE SSI-tuned dataset as VAE SSI to differentiate it from the initial VAE dataset.
 
\subsection{Performance evaluation}
The performance of the baseline models was assessed under two distinct settings. The first setting evaluated the models' proficiency in accepting genuine signatures, while the second focused on rejecting non-genuine signatures. Performance was quantified using False Reject Rates (FRR), probabilistically complementary to True Accept Rates (TAR). In contrast, the models' ability to reject non-genuine signatures was measured using False Accept Rates (FAR), probabilistically complementary to True Reject Rates (TRR), across the four attack scenarios described above. For convenience, the Half Total Error Rate (HTER), the average of FAR and FRR, is also reported for easier comparison between different systems.

\section{Results and discussion}
\label{ResultsAndDiscussion}
\subsection{Baseline performance} 
\label{baseline_perf}
\begin{figure}[ht]
  \centering
  \includegraphics[width=2.6in, height =1.8in]{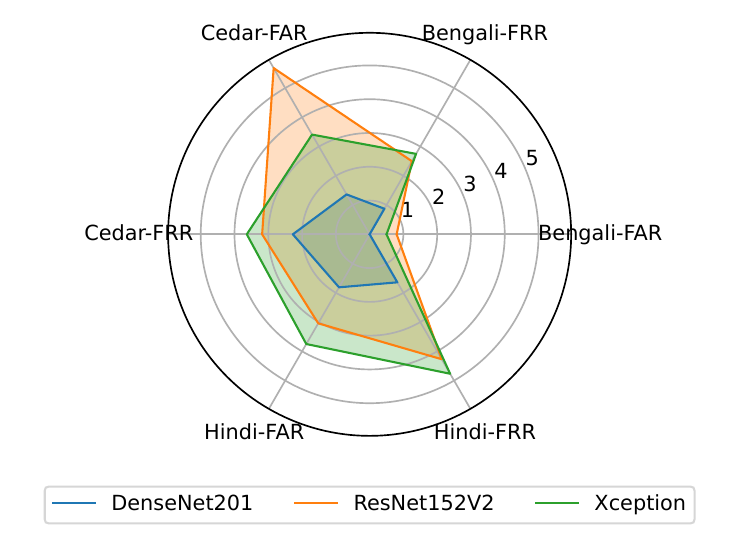}
  \caption{Radar chart illustrating the performance of three baseline models—DenseNet201, ResNet152V2, and Xception—on the CEDAR, BHSig260-Bengali (Bengali), and BHSig260-Hindi (Hindi) datasets. The axes represent each dataset's False Accept Rates (FAR) and False Reject Rates (FRR). All models achieved error rates below 6\% for both FAR and FRR. DenseNet201 performed best with FARs and FRRs lower than 2\% across datasets, followed by ResNet152V2 and Xception.}
  \label{baseline}
\end{figure}

Figure \ref{baseline} presents the performance of the three baseline models, namely Xception, ResNet152V2, and DenseNet201, across the CEDAR, BHSig260-Bengali, and BHSig260-Hindi datasets. All three models consistently achieved error rates below 6\% in both FAR and FRR. DenseNet201 was the top performer with an HTER of 1.81\%, followed by Xception (3.52\%) and ResNet152V2 (4.43\%) for the CEDAR dataset. For BHSig260-Bengali, DenseNet201 achieved an HTER of 0.44\%, followed by ResNet152V2 (1.65\%) and Xception (1.65\%). For BHSig260-Hindi, DenseNet201, ResNet152V2, and Xception obtained HTERs of 1.73\%, 3.66\%, and 4.26\%, respectively. The performance closely followed the results presented in the literature \cite{Inception2015, DeepSign2021AllBaselines}.

\begin{figure}[htp]
  \centering
  \includegraphics[width=3.5in, height= 1.56in]{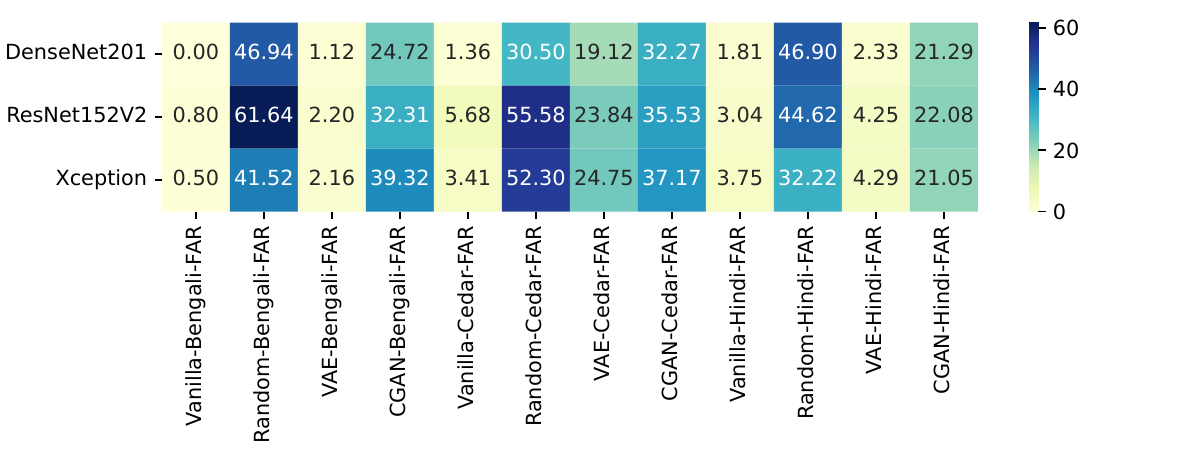}
  \caption{Attack performance on the baseline ASVs. Synthetic signatures were generated using skilled forgeries as a reference, and each DGM ran for $800$ epochs. The higher the SSIM, the more similar the generated signature is to the forgeries, resulting in a lower FAR. Conversely, the lower the SSIM, the farther the generated signature is from the reference forgeries, resulting in a higher FAR. \textit{Random} signature attacks resulted in high vulnerability, with FARs over 30\% on average. \textit{VAE} attacks produced FARs closest to original forgeries, except for the CEDAR models, which had higher FARs around 20-25\%. \textit{CGAN} attacks caused the highest FARs, exceeding 29\% on average across all models and datasets.}
  \vspace{-0.1in}
  \label{attack-baseline}
\end{figure}

\begin{figure*}[htp]
  \centering
  \includegraphics[width=6.7in, height = 1.85in]{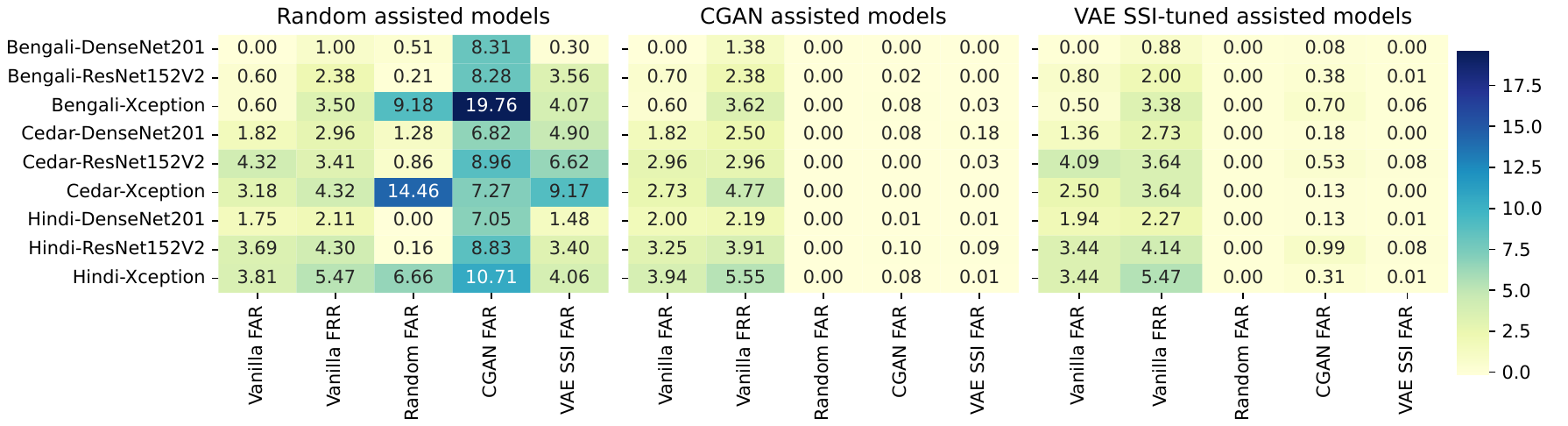}
  \caption{Heatmaps showing the performance of retrained models under evaluated attacks. The x-axis labels indicate the type of attacks, and the y-axis labels denote the dataset and underlying deep learning architecture. Heatmap values indicate FAR for each combination of rows and columns. The first heatmap (first five columns) represents models retrained with random signature images. The second and third heatmaps (middle and last five columns) correspond to models retrained with CGAN-generated and VAE-SSIM-tuned-generated signatures, respectively. CGAN and VAE-SSIM-based retraining reduced FARs below 1\%, demonstrating their superiority over Vanilla or random signature-assisted re-training. The Random, CGAN, and VAE-SSIM attacks impact only FARs, so only FARs are reported for these. In contrast, the Vanilla attack scenario included both FAR and FRR. Comparing the Vanilla FRR in these heatmaps with the FRRs in Figure \ref{baseline} shows that the retrained models maintain or closely match the FRR of the initial models, significantly reducing the FARs.}
  \label{retrained-models}
\end{figure*}

\subsection{Attack performance} \label{Attack_Per}
The effect of described attack scenarios is presented in Figure \ref{attack-baseline} measured using FAR. 

\textbf{Random attacks} Figure \ref{attack-baseline} reveals that although our baseline systems exhibited low FARs against skilled forgeries, they were highly vulnerable to even random attacks across all three models. In particular, ResNet152V2 trained on the Bengali dataset appears to have the highest FAR on random attacks. Similarly, on average, other models' FAR on random attacks reached higher than 30.5\%. 

\textbf{CGAN-assisted attacks}
CGAN-generated signatures with lower SSIM caused substantial damage to the baseline ASVs when used as attack samples. Specifically, the CGAN-based attack achieved a $29.5$\% FAR across the ASVs and datasets. This error rate far exceeded that of original human forgeries. The reason is that the CGAN-generated signatures were far from skilled forgeries and presumably closer to genuine signatures, resulting in higher FARs. In contrast, the VAE-generated signatures were closer to the human forgeries, resulting in rejections by the ASVs and causing lower FARs than the CGAN-generated signatures.

\textbf{VAE-assisted attacks}
Presented in Figure \ref{attack-baseline}, the attack results using VAE datasets closely mirrored our anticipated outcomes. We observed that most of the results from VAE-generated signatures came close to the original forgery samples across all datasets, with a slight exception for those trained on CEDAR. Specifically, DenseNet201, ResNet152V2, and Xception trained on Vanilla Bengali had FARs of $0\%, 0.8\%$, and $0.5\%$ for Vanilla attacks, compared to $1.12\%, 2.20\%$, and $2.16\%$ for VAE attacks, respectively. Similarly, for the models trained on Vanilla Hindi, the FARs for Vanilla attacks were $1.36\%, 5.68\%$, and $3.41\%$, compared to $2.33\%, 4.25\%$, and $4.29\%$ for VAE attacks. CEDAR-trained models were more affected by this type of attack, with significantly higher FARs of $19.12\%, 23.84\%$, and $24.75\%$ resulting from the VAE attacks.

There exists several other generation methods as listed in \cite{DeepGenerativeSurvey2022}. We posit that different synthetic signature generation methods will have varying levels of attack and robustness impact.  

\subsection{Countermeasure and performance} \label{retrain}
\textbf{Random-assisted models}
The first heatmap (five columns from the left) in Figure \ref{retrained-models} suggests that while the FAR and FRR tested on Human Genuine and Human Forgeries experienced slight changes, random-assisted models demonstrated higher robustness to random and generated attacks (compare the FARs in Figure \ref{attack-baseline} and Figure \ref{retrained-models}). 


\textbf{CGAN-assisted models}
The second heatmap (five columns in the middle) in Figure \ref{retrained-models} suggests that the CGAN-assisted models showed high robustness against each attack. In particular, the FAR for all random attacks has been minimized to 0\% and CGAN attacks under 0.10\%, followed by the Vanilla attack under 0.18\% in the worst case. 

\textbf{VAE-assisted models}
The third heatmap (five columns on the right) shown in Figure \ref{retrained-models} suggests that the VAE-assisted retraining drastically improved the robustness of the models under all the attacks. In particular, VAE-assisted models also obtained $0\%$ FAR on random attack. The models also show significant improvements in CGAN attacks, despite not as much as CGAN-assisted models and VAE-SSI attacks. Particularly, the FAR falls in the range of 0.08 - 0.99\% for CGAN attacks and 0.0-0.08\% for VAE-SSI attacks. The models' FAR on VAE-SSI attacks was slightly lower, with an average of $0.026\%$, compared to $0.038\%$ for CGAN-assisted models.

The results support the hypothesis presented in \cite{Zhao2020DataInjection} that mixing up synthetic signatures farther away from the human forgeries (negative samples) increases the variance of the impostor class and, in turn, reduces the false accept rates. 

\subsection{Implications} \label{result_app}
The research revealed severe vulnerabilities in data-driven ASVs to random and deep generative attacks. In other words, the existing data-driven ASVs needed to have detected state-of-the-art generative forgeries to an alarming extent. The proposed countermeasure, i.e., retraining the ASVs using SSIM-tuned generated signatures, effectively reduced the impact of zero-effort, random, VAE, and CGAN-assisted attacks. This research thus emphasizes the continuous study of emerging threats to ASVs and possible countermeasures, enabling the widespread adoption of data-driven automatic signature verification and their trustworthiness among users.

\section{Conclusion and future work}
\label{ConclusionFutureWork}
This study evaluated the vulnerabilities of data-driven offline ASVs to random and generative attacks and proposed effective countermeasures. Two advanced deep generative models, Variational Autoencoders (VAEs) and Conditional Generative Adversarial Networks (CGANs), were used to generate synthetic signatures that challenged baseline ASVs. The study demonstrated that CGAN-generated signatures, with lower SSIM, caused significant FARs, highlighting a critical security risk. The findings revealed a strong negative correlation between SSIM and FAR, indicating that lower-quality synthetic forgeries are more likely to be accepted as genuine signatures. This insight led to the development of a novel countermeasure involving SSIM-tuned retraining. Incorporating synthetic forgeries with controlled SSIM into the training process significantly improved the robustness of ASVs against various attack scenarios.

The retrained models, particularly those using VAE-SSIM-tuned signatures, exhibited FARs below 1\% under CGAN and VAE-SSI attacks, demonstrating the effectiveness of our approach. This study underscores the importance of continuously evaluating and enhancing ASVs to mitigate evolving threats. In the future, we plan to investigate additional image quality metrics, such as the Fréchet Inception Distance (FID) and Universal Image Quality Index (UIQI), to better understand their relationship with attack success rates. Additionally, we aim to explore the potential of using genuine human signature datasets to improve system robustness further. Finally, extending our research to online signature verification and writer-independent models will be crucial for developing comprehensive and resilient ASVs.

\section{Acknowledgment} We thank the anonymous reviewers for their invaluable feedback and comments, which significantly improved this manuscript. We are also grateful to The Bucknell Program for Undergraduate Research (PUR) for supporting An Ngo.

{\small
\bibliography{references}
}

\end{document}